\documentclass[runningheads]{llncs}
\usepackage{graphicx}
\usepackage{subfig}
\usepackage{hyperref}
\usepackage{todonotes}

\newcommand{\varpos}{\mathrm{Var}_\mathrm{P}}

\begin{document}
\title{Rhoban Football Club: RoboCup Humanoid KidSize 2019 Champion Team Paper}

\titlerunning{Rhoban Football Club: RoboCup Humanoid KidSize 2019 Champion}
%
\author{
  Lo\"{i}c Gondry \and
  Ludovic Hofer \and
  Patxi Laborde-Zubieta \and
  Olivier Ly \and
  Lucie Math\'e
  Gr\'egoire Passault \and
  Antoine Pirrone \and
  Antun Skuric
}
\authorrunning{Rhoban Football Club}
\institute{
    Rhoban Football Club Team,\\
    LaBRI, University of Bordeaux, France\\
    \email{team@rhoban.com}\\
    Corresponding author: Patxi Laborde-Zubieta\\
    \email{patxi.laborde-zubieta@u-bordeaux.fr}
}
\maketitle              
\begin{abstract}
  In 2019, Rhoban Football Club reached the first place of the KidSize soccer
  competition for the fourth time and performed the first in-game throw-in in
  the history of the Humanoid league.
  Building on our existing code-base, we improved some specific functionalities,
  introduced new behaviors and experimented with original methods for labeling
  videos.
  This paper presents and reviews our latest changes to both software and
  hardware, highlighting the lessons learned during RoboCup.
\end{abstract}
\section{Introduction}
This article presents some of the elements which led to the fourth consecutive
victory of our team in the RoboCup KidSize Humanoid league.
We also obtained the first place in the drop-in tournament for the third time in
a row and the first place at the technical challenges competition for the first
time.
Our robots scored 30 goals, received 11 goals and performed the first in-game
throw-in in the history of the Humanoid league.

This year we mainly pursued three objectives: moving toward more dynamic
gameplay, improving our performance at the technical challenge competition
and reducing the complexity of our code base which has grown each year since we
first participated in 2011.
While introducing several new functionalities, we still managed to reduce the total
number of lines of code used for the competition from 196,000 to 163,000.
All the code and configuration file we used during the competition are
available\footnote{\url{https://www.github.com/Rhoban/workspace/releases/tag/public_2019}}
along with some documentation.

The structure of the paper is as follows: section~\ref{sec:model} introduces
the tools used to model our robot, recent hardware changes are presented in
section~\ref{sec:hardware}, new motions and improvements are detailed in
section~\ref{sec:motion}, our ongoing work regarding perception and data
acquisition is described in section~\ref{sec:perception} and finally the dynamical aspect
of our strategy is presented in section~\ref{sec:strategy}.

\section{\label{sec:model}Model}
The 3D model of the robots used during Robocup 2019 can be
accessed\footnote{\url{https://cad.onshape.com/documents/f3bdef32bffd81536fce83d1/v/779c691df8f135bba01eead1/e/a530b1889ee09acb5e1d7ff9}}
in the online CAD software
\textit{OnShape}\footnote{\url{https://www.onshape.com/}}.

\subsection{CAD to standard model (URDF/SDF)}

Several CAD tools are commonly used to design robotics parts\footnote{Famous examples
are \textit{Dassault Solidworks}, \textit{Autodesk Inventor} and \textit{Catia}},
mostly based on constraints geometry design. On the other hand, standard robot description
format emerged, notably URDF and SDF\footnote{\url{http://sdformat.org/spec}}, driven by the
ROS community\cite{quigley2009ros}. They are XML files describing the robot
architecture, including transformation matrices, information about dynamics (mass,
center of mass, inertia), collisions and visualisation geometry.
In order to manufacture them, we used to design our robot with such CAD tools.
But even if the model carried all the information needed for the standard robot
description, the description was produced with separate tools. As a result, we could not ensure the
consistency between the CAD and the description model.
For that reason, we switched to \textit{OnShape}, an emerging
CAD software that includes an API allowing to request information about
the 3D model.
It allowed us to develop \textit{onshape-to-robot}\footnote{\url{https://github.com/rhoban/onshape-to-robot/}},
a tool that seamlessly produces an URDF from a CAD model that can be
used without any change for all our applications.

\subsection{Using model in online code}
In order to compute frame transformations online using the robot , we developed a library on top of
\textit{RBDL}\cite{felis2017rbdl} to load the \textit{URDF} and request it.
\textit{onshape-to-robot} allows you to attach manually frames in your model, directly in
the CAD design-time, that appear in the final robot description and allow you to
compute transformation matrices using the DOFs of your robot.

\subsection{Physics simulation}

The robot description model that is produced this way can also be used for physics simulation,
like \textit{Bullet}\footnote{\url{https://github.com/bulletphysics/bullet3}}.
However, simulating the collisions of the actual parts is computationally expensive,
first because the exported parts are represented by unstructured triangular surfaces, but also
because of all the small details (like screw holes) which are not relevant for our use case.

\begin{figure}
\begin{minipage}{.49\textwidth}
  \centering
  \includegraphics[height=3.5cm]{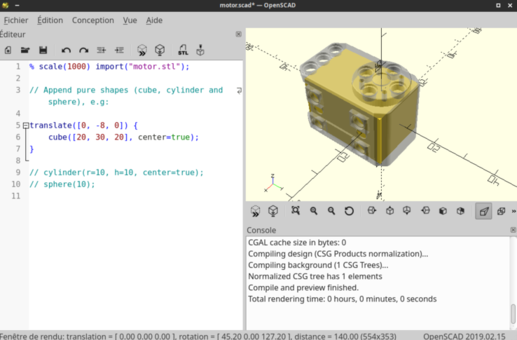}
  \captionof{figure}{Semi-automatic system to approximate 3D model with shape representation.}
  \label{fig:pureshape}
\end{minipage}%
\hfill
\begin{minipage}{.49\textwidth}
  \centering
  \includegraphics[height=3.5cm]{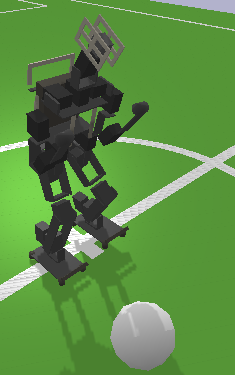}
  \captionof{figure}{Sigmaban approximated with 3D shapes in PyBullet physics simulator.}
  \label{fig:pybullet}
\end{minipage}
\end{figure}

To tackle this issue, we introduced a semi-automatic system that allows to approximate
those complex 3D shapes with pure geometry: union of cubes, spheres and cylinders (Fig.~\ref{fig:pureshape}).
This also allows to ignore some small parts that are not useful in the collision
world. We are then able to simulate a physics model of our robot (Fig.~\ref{fig:pybullet}). Even if the
discrepancy between simulation and real world is high, motions like walking, kicking and standing up
can be reproduced, allowing to do some motor test before porting it to the real robot.

\section{\label{sec:hardware}Hardware}
We made only few changes on the hardware. We added some piano wire arcs to
protect from falls, we switched to a four cells battery and we changed the shape
of the feet and the hands. The last two improvements are described in
section~\ref{sec:motion}.
\subsection{Protections}
One of the challenges in our league is that robots should be able to withstand
falls. For example, our robots can fall up to 20 times during a game. It
regularly resulted in the breaking of a motor of the neck. After several
attempts, it was clear to us that a software safety on its own was not enough.
Hence, to absorb a part of the impact we added 3mm thick piano wire arcs at the
front and at the back of the robot. By doing so,
in 2018 and 2019 we had a significant decrease in the number of broken motors in
the neck.
But it is still not reliable enough as the many shoulder motors that we
broke can tell.

To improve the protection of our robots, we would like to try other materials
such as spring steel strips, which are less likely to bend perpendicularly to the
impact, or dense foam. In order to assess rigorously the effectiveness of
different solutions, the impacts when falling should be measured. For example, using 
motion capture or force platforms. Another way to protect the motors is to use
clutches, but for the moment we did not find any solution fulfilling the
requirements in term of weight and space used.

As mentioned during the second edition of the workshop ``Humanoid Robot Falling:
Fall Detection, Damage Prevention and Recovery
Actions''\footnote{\url{https://iros2018wsfallingrobots.wordpress.com/}}, this
problem is a research topic of growing importance. And the robots of the
KidSize league happens to be an interesting benchmarking environment for this subject:
solutions can be tested under realistic conditions during matches and it is much easier to
safely experiment with smaller humanoid robots.

\subsection{Battery}
In our design, the 6 motors of a leg are connected in series. We observed
voltage drop up to 4 volts between the first and the last motor of the leg
during dynamic motions. When using three cells batteries, this leads to a voltage
around 8 Volts in the ankle. While the four cells batteries are out of the
specified range for the dynamixel motors since they deliver 16V, they strongly increase the available
torque and reduces the ohmic power loss. In the the past, we were using MX-64
motors in the legs and a lower position of the center of mass during the walk
motion, this led to frequent overheating when using four cells batteries. Now
that we have MX-106 motors and a smoother walking engine, we can safely use four
cells batteries without risking overheating. Increasing the voltage was one of
the key elements to obtain a more powerful kick. When the ball was properly
positioned, our robot managed to kick more than seven meters.
\section{\label{sec:motion}Motions}
\subsection{New walk engine}

\begin{figure}[h]
  \centering
  \includegraphics[scale = 0.2]{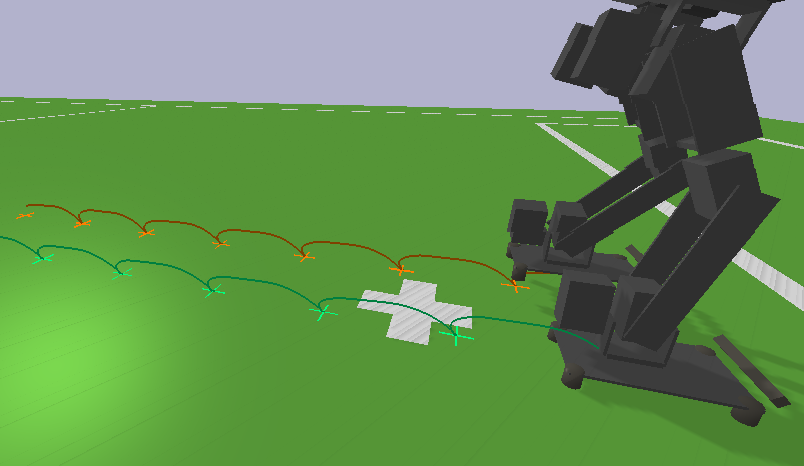}
  \caption{Trajectory of the feet during walk engine (physics simulation).}
  \label{fig:walkTraj}
\end{figure}

We designed a new walk engine, still based on single support, with the goal of reducing the complexity of the
former one that included too many unused parameters, making it less maintainable.
We now use cubic splines to represent the trajectories of the feet
(see Fig.~\ref{fig:walkTraj}). They allow us to control both position
and speed at specific knots.
Trajectories are updated only at each support foot swap and described in the
trunk frame.

We wanted the foot to reach its nominal speed before
touching the ground and to decelerate after leaving the ground.
This way, the foot in contact with the ground would have a constant speed
which should lead to a steady and continuous speed of the trunk. However, the robot is
performing much better when the foot touches and leaves the ground with vanishing speed.
One hypothesis is that it is better to touch the ground with no speed
for stability reasons. Moreover, the exact time when the foot touches the ground can
vary from one step to one another.

The walk engine code can be found on our open source repository under \texttt{Motion/engines/}
directory.

\subsection{Throw-in}

This year, the throw-in rule has been introduced.
While it is currently allowed for robots to perform it by kicking,
we decided to move directly to human-like throw-in performed with the hands.
The motion for the throw-in is created the same way as other kicks: we define
splines associating time points with angular targets for the motors.
The target between specified time points is obtained through linear interpolation.

We designed the throw-in motion using splines with several keypoints, see Fig.~\ref{fig:ti}:
initialization, bending the knees, unfolding the arms, leaning forward, holding
the ball, straightening, lifting the arms and unfolding the knees, moving the
arms behind the head, throwing the ball and back at initialization.

\newcommand{\throwIn}[1]{\includegraphics[width=.16\textwidth]{#1.png}}

\begin{figure}[h!]
    \centering
    \throwIn{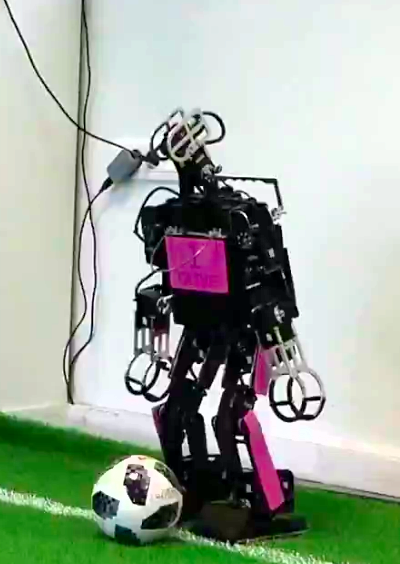}
    \hfill
    \throwIn{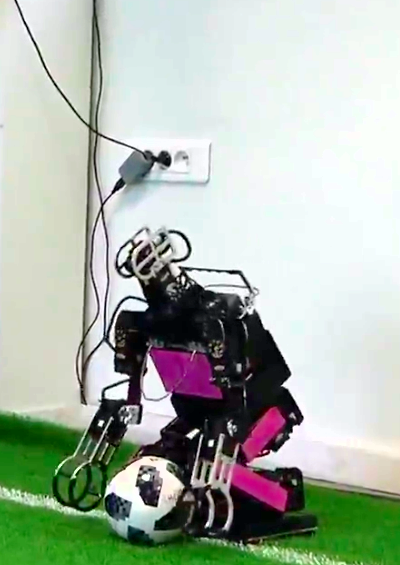}
    \hfill
    \throwIn{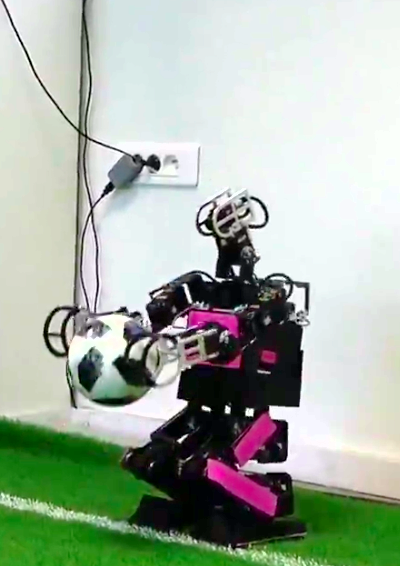}
    \hfill
    \throwIn{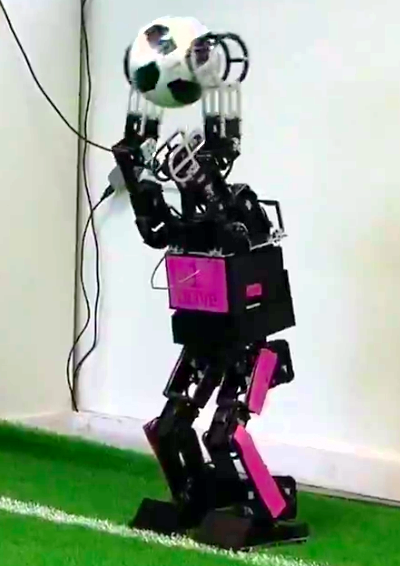}
    \hfill
    \throwIn{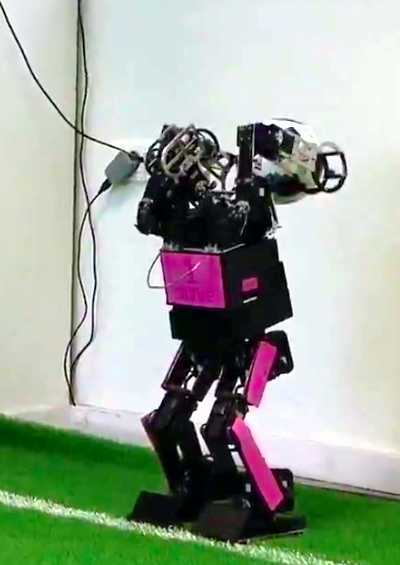}
    \hfill
    \throwIn{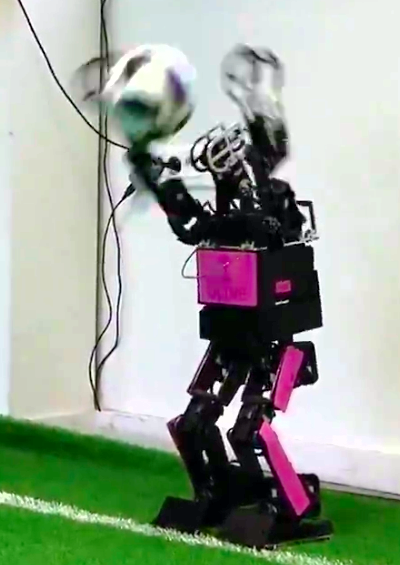}
    \caption{Differents steps of the Throw-in.}
    \label{fig:ti}
\end{figure}

While simulation using \emph{PyBullet} was not accurate, it allowed to design a
coarse approximation of the motion before fine tuning the targets on the robot.

One of the main challenge to design the throw-in was to handle the fact that our
arms have only three degrees of freedom.
While the mechanical design is entirely sufficient for standing-up,
it does not allow to control the distance between the hands when the elbow are bent.
Therefore, amplitude of the motion on the elbow was limited in order to
maintain grasp on the ball.

\begin{figure}[h]
  \centering
  \includegraphics[scale = 0.13]{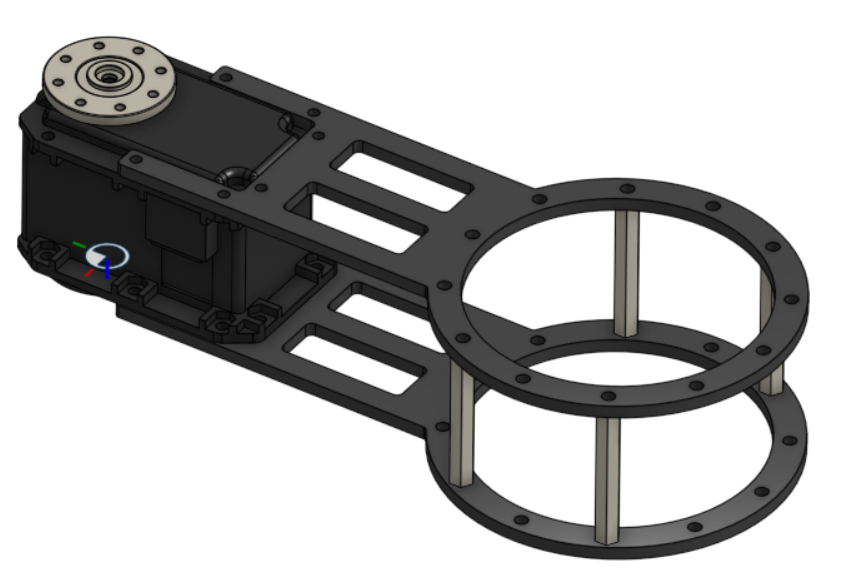}
  \caption{Assembly of one hand of the robot in \textit{OnShape}.}
  \label{fig:hand}
\end{figure}

In order to lift the ball more easily, we designed new hands for the robot, see Fig.~\ref{fig:hand}.
The main idea was to create two metal plates in each side of the elbow motor and join
them with spacer screws. On one side, the plate has a notch for the motor and, on the
other side, there is large hole to surround the ball.

While designing the motion, we managed to have the ball bouncing and rolling up
to $4.5$ meters.
Since the field is only 6 meters wide, we had to slow down the speed during
the throw-in phase in order to have a more appropriate length for passes.

Next year, we plan to improve the throw-in so that the robot can adapt the direction
by changing the orientation of the torso while the ball is in the air.

\subsection{High Kick}

One of the key aspect to lift the ball while kicking is the point in contact
with the ball at impact.
We refer to this part by the name of \emph{kicker}.
Previously, we used a \emph{kicker} designed to give a rotational impulsion to
ball so that it could roll on the grass.
This year, in order to perform in the High Kick technical challenge, we designed
new \emph{kickers}.

To lift the ball, the contact point at impact has to be below the center of
mass of the ball.
Since we use high studs to stabilize on artificial grass and the ball has a
radius of only 7.5 centimeters, the margin for the motion is relatively small.
We decided to separate the new \emph{kicker} in two different parts:
the first one is thin and raises the ball which rolls over it,
the second one hits the ball after it has left the ground making it easier to
hit below the center of gravity.

\newcommand{\highKick}[1]{\includegraphics[width=.32\textwidth]{#1.png}}

\begin{figure}[h!]
    \centering
    \highKick{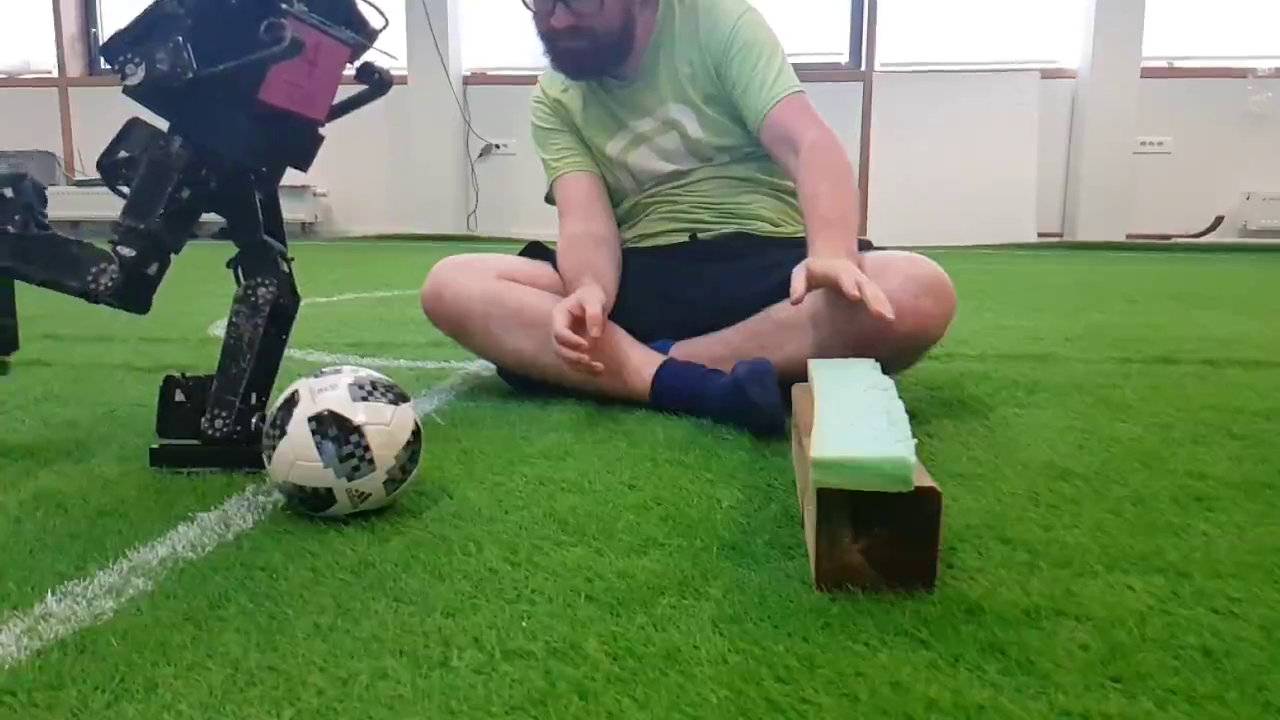}
    \hfill
    \highKick{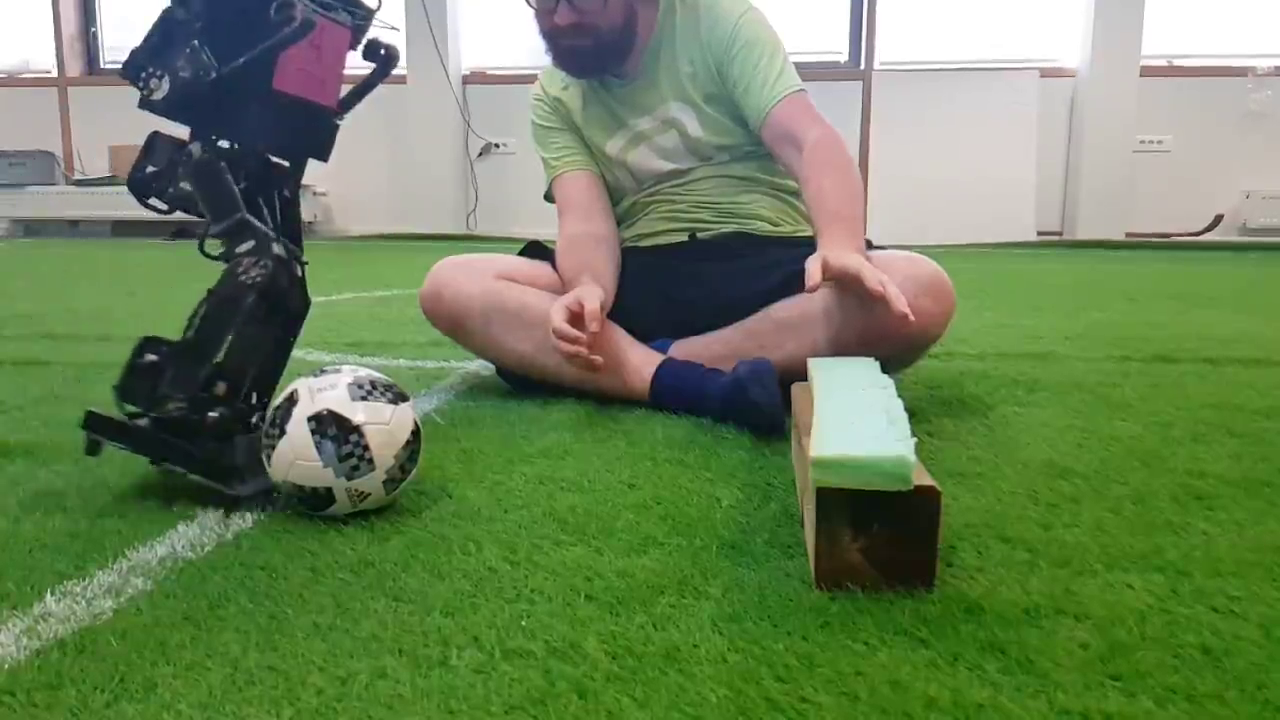}
    \hfill
    \highKick{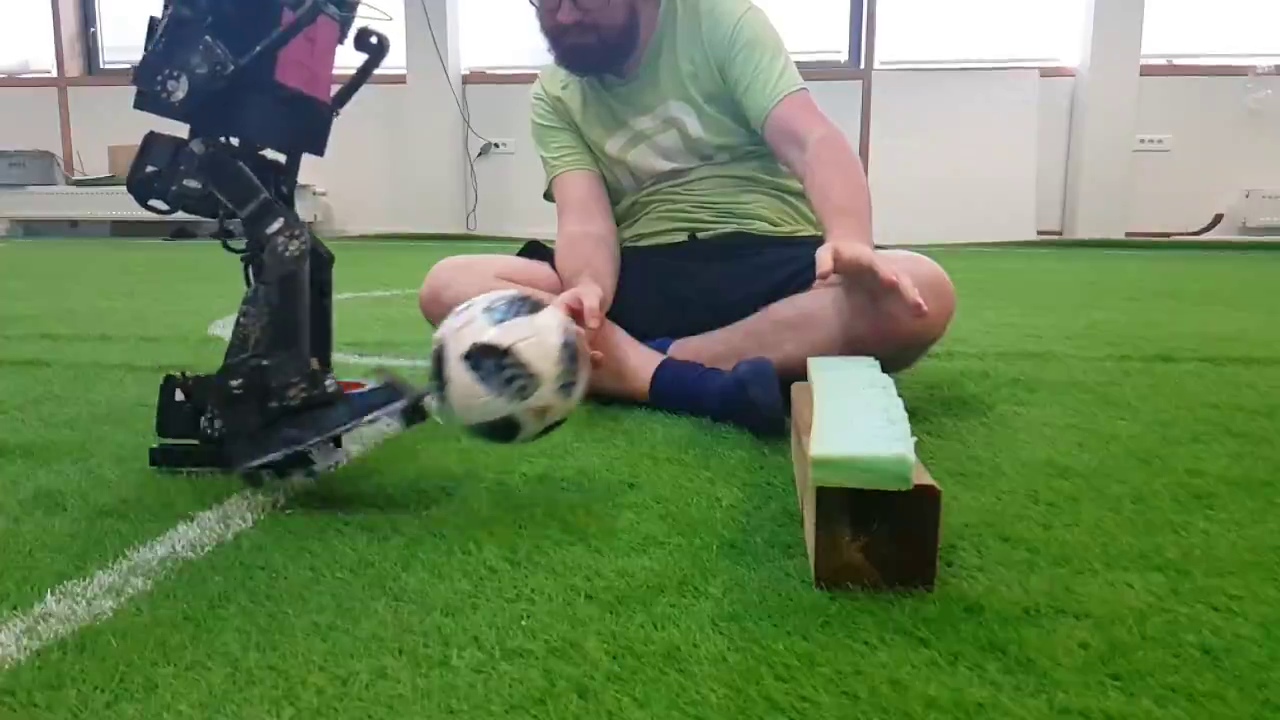}
    \\
    \highKick{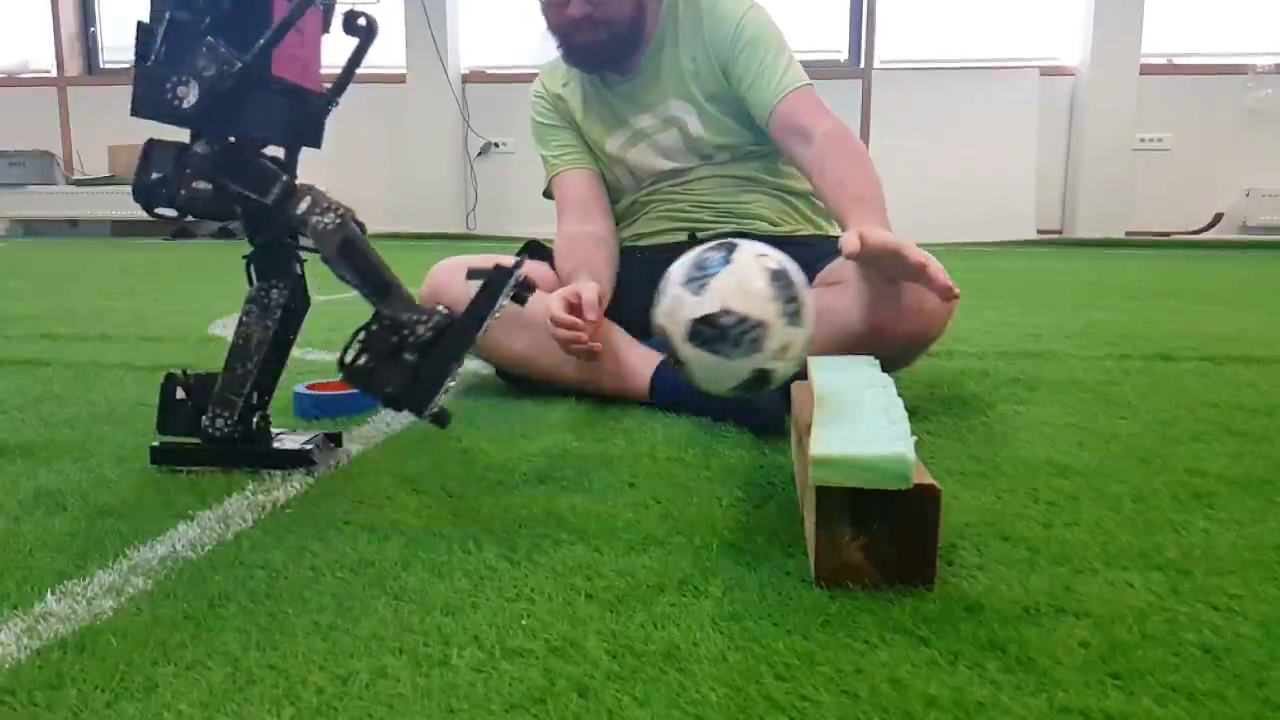}
    \hfill
    \highKick{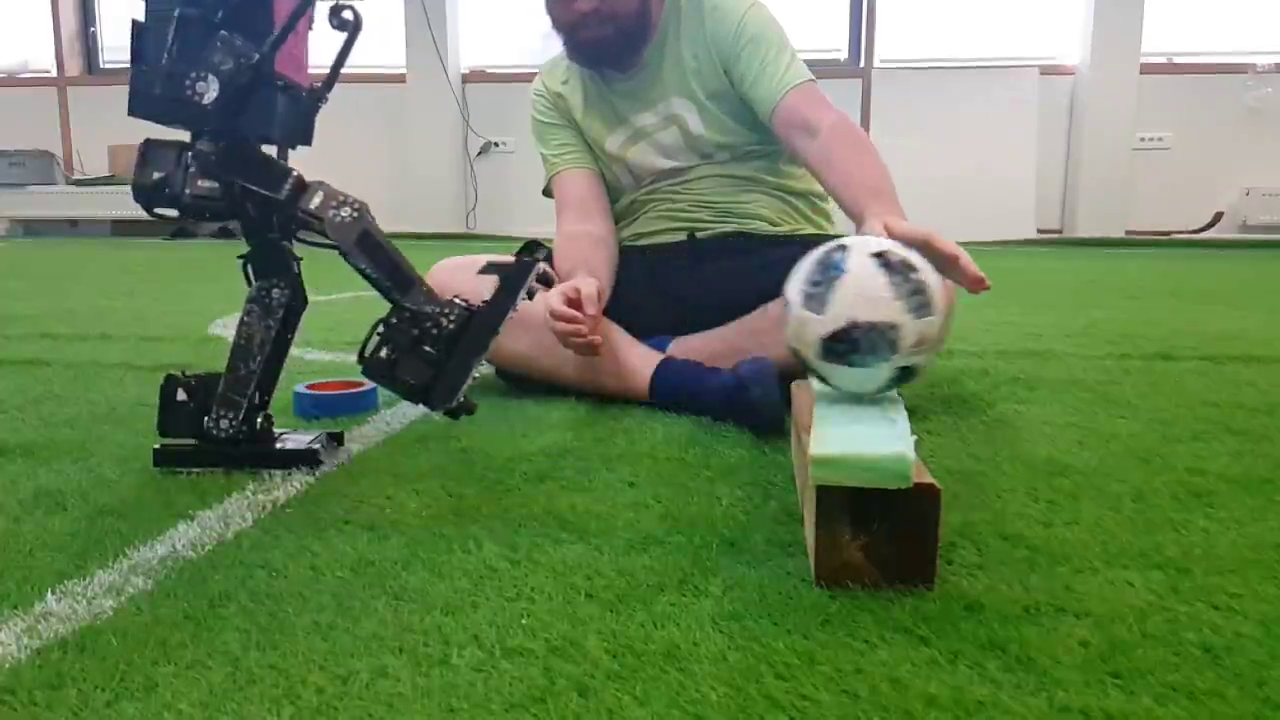}
    \hfill
    \highKick{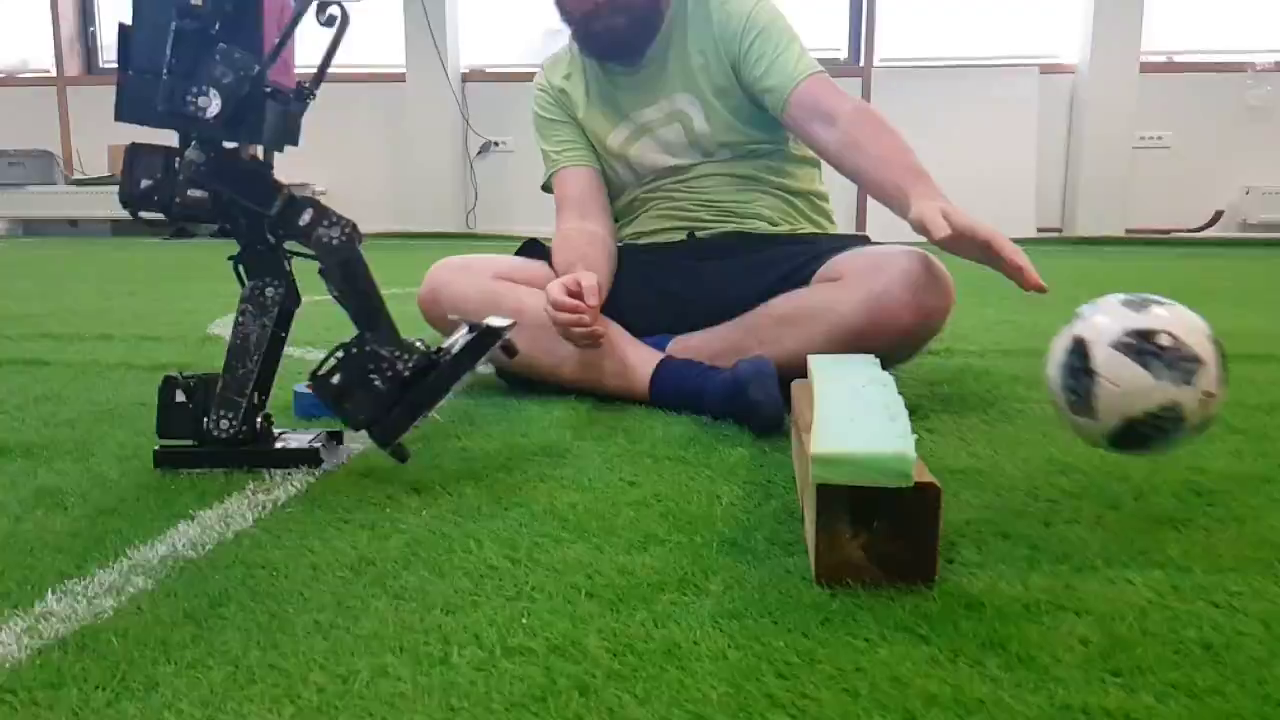}
    \caption{Differents steps of the High kick.}
    \label{fig:hk}
  \end{figure}

This new kicker allowed us to outperform the other teams during the technical
challenge, our robot scored over a bar of 20cm, while the second best
performance in our league was achieved by the CIT Brains who kicked the ball
above 12cm. Fig.~\ref{fig:hk} shows the steps of this motion.
An interesting fact about this high kick is that one of our robot did kick above
another fallen robot during the quarter finals.
Although not intentional, this kick is definitely a step toward the use of a
third dimension in the RoboCup Humanoid league.

\section{\label{sec:perception}Perception and Localisation}
This year, we aimed high regarding the modifications of our perception and
localisation modules.
Unfortunately, our system for labeling videos still lacked some robustness and
the training procedure for our neural networks included a few bugs which
impacted negatively its results during the competition.
This sections presents our promising development of these modules along with
preliminary conclusions based on their use during RoboCup 2019.
\subsection{Labeling videos}
Most of the teams in the RoboCup Humanoid league now uses neural network to detect or
classify features in their perception module. 
In order for the module to work on-site, it is generally required to manually
label large datasets of images acquired on-site.
The labeling of images requires a significant amount of human time and adding
new features to detect for the robots increases the time spent labeling.
We also previously used manual tagging, with the help of a collaborative
on-line tool we developed for that purpose \footnote{\url{https://github.com/rhoban/tagger}}.

This year, we decided to take a paradigm shift, moving from labeling images to
labeling videos.
The main idea is quite simple, if we can retrieve the pose and orientation of
the robots camera inside the field referential for each frame,
then the position of field landmarks inside images can easily be obtained.
Moreover, by synchronizing the video streams from multiple robots,
it is possible to share annotations among them. 

Accurate estimation of the orientation of the cameras is a difficult problem,
in order to tackle it, we experimented two different methods:
using ViveTrackers on the head of the robots and combining manual labeling with
odometry.
Both methods share similar issues regarding time synchronization between devices
and its impact on orientation estimation.
We used two complementary schemes to reduce the uncertainty on camera
orientation.
\begin{enumerate}
\item We used specific tool (\verb!chrony!\footnote{\url{https://chrony.tuxfamily.org/}})
  to synchronize all the information streams.
\item The impact of timing differences is mitigated using the following
  acquisition method for videos.
  Robots alternate between two different phases:
  walking to a randomly generated location and slowly scanning the environment
  while standing still.
  During extraction of labels, only the images obtained during the scanning
  phase are considered.
  This proved to be necessary because the head of the robot is shaking when the
  robot is walking, thus increasing the uncertainty on the orientation of the
  camera.
\end{enumerate}

Access to the fields for data acquisition is a scarce ressource during the setup
days.
In order to optimize the usage of the time we were given,
we created a specific training scenario in which the field is separated in as
many zones as the number of robots used for acquisition.
\begin{itemize}
\item A zone is allocated to each robot. 
\item During 2 minutes, each robot alternates between moving to a random
  location inside its zone and scanning its environment.
\item Zones are separated by a safety buffer of around 50cm to reduce the risk
  of collision between robots.
\item During the training, perception is disabled and robot relies solely on
  odometry to estimate its position.
\end{itemize}

\subsubsection{Automatically through Vive}

\textit{Vive} is an indoor tracking system developed by \textit{HTC}. It is
based on active laser emitters base stations called \textit{Lighthouses} that
sends laser sweeps on a known frequency. Infrared receivers are used and the
time when they are hit by the sweeps of lighthouses is used to compute the
position of the object they are attached to. There are two generation of the
lighthouses. The first generation only works by pairs and covers a maximal tracking
area of $5m\times5m$. With the second version, the area can be increased up to
$10m\times10m$ by using four lighthouses. Initially designed to track \textit{Vive} controllers
and helmets for virtual reality application, it is now also possible to buy
simple trackers that you can attach to anything to track its position.

This makes ground truth possible with attaching a \textit{Vive} tracker to the head of
the robot, and capture some logs. The only thing that is required is then the transformation
from the tracker frame to the camera frame, and the ability to project a known object
3D position onto image.

\textit{Vive} trackers and lighthouses are easy to carry and deploy.
Moreover, the calibration phase before being able to track objects is fast. This
makes it a better choice than motion capture for on-site calibration, while
being more affordable and with a wider working area. We used it during the
\textit{German Open 2019 Humanoid KidSize} competition to generate
automatically labelled images.

A calibration phase is still needed to find the 3D transformation from the lighthouses
to the field frame, and also to tag the balls position on the field every time we move
them. To achieve that conveniently, we use a \textit{Vive} controller which is itself tracked and
equipped with trigger button to mark some known position on the field and find the optimal
3D transformations, or show the balls position at the beginning of a log. We developed a
custom tool to do that\footnote{\url{https://github.com/Rhoban/vive_provider}} on the
top of OpenVR SDK\footnote{\url{https://github.com/ValveSoftware/openvr}}.

Even if most of the data obtained this way was suitable for training our neural networks,
the accuracy of such technology can be discussed\cite{niehorster2017accuracy}.
In order to have a better control on the calibration and the data fusion algorithm
used to comptute the position using the IMU and light beam datas,
the authors of
LibSurvive\footnote{\url{https://github.com/cnlohr/libsurvive}}
reverse engineered \textit{Vive}.
Having access to the low level data allowed better
positioning results\cite{borges2018htc}. The support for
the second lighthouse generation is currently being completed.

One of the drawbacks intrinsic to this method is that it cannot be used during real games.

\subsubsection{Combining labeling and odometry}
Finding a camera pose from 3D-2D correspondances is a well-known
problem\cite{Lepetit2009},
by labeling the position of keypoints of the field in an image,
it is relatively easy to retrieve an accurate estimation of the pose of the
camera which took the image.
However, it is not realistic to apply this method for every image of a video
for two majors reasons: this would require large amount of human-time to label
a video and some frames do not even contain keypoints from the field.

Using odometry to extrapolate the pose of the camera before and after a labeled
frame helps to strongly reduce the labeling burden.
Experimentally, we noticed that labeling around 10 frames for a 2 minute
session containing more than 1500 usable frames hold satisfying results.

Currently, the major flaw of this method is that it requires the robots to stop
walking and reduce the scanning speed in order to improve the pose estimation. 
Those conditions are rarely met during real games, making it difficult to
extract data from real games, a crucial point to enhance opponent detection.
In the future, we hope to take leverage on visual-inertial odometry methods\cite{Gui2015} to
enhance the accuracy of pose estimation in more dynamic situations.
This would allow to easily label videos of entire games quickly, thus tackling
the problem of building large datasets for neural network training.

\subsection{Multi-class ROI and classification}
Last year, our perception system was mainly based on three specific pipelines,
one for detecting the ball, a second one detecting the base of the goal posts
and a third one to detect the corner of the arena field.
The two first systems were roughly similar, but each system had its own neural
network and some specificities to identify region of interests\cite{n2018rhoban}.
In order to make the perception system simpler while covering more type of
features, we decided to use a single system to identify the region of interests
and classify the type of feature.
While less accurate for centering the features in the region of interests,
it allows to incorporate additional features such as line intersections or
opponents more easily.

While this new system yielded promising results during the preparation of the
RoboCup, we struggled to obtain decent results on-site and were forced to
limit the perception to two different type of features: balls and base of goal
posts.
We initially thought that the main problem was the fact that the posts
were thinner than the posts we used in our laboratory.
However, after RoboCup, we ran a thorough code review and found 3 major bugs
between the training process of the neural networks and the online prediction of
the class.
Once we solved them, we were able to include more classes while having an
accuracy rate much higher than what we had during RoboCup.
Due to these bugs, it is not possible to provide meaningful results for the code
we used during RoboCup.
However, the final results promise a strongly improved perception system for
next year.

\subsection{Localisation}
We use a three-dimensional particle filter for localization which includes the
position and the orientation of the robot on the field.
It fuses information from both: the perception module and the odometry.
Due to the major issues in perception at the beginning of the competition,
we decided to strongly reduce the exploration for the first games.
Increasing the confidence in the odometry allowed us to stay into the
competition until we improved the perception part.

The position and the orientation of the robot used to be defined respectively as
the average of the positions and the orientations of the particles. The major
change we introduced this year was to fit a Gaussian mixture model on the set of
particles by using the Expectation-Maximization\cite{Bilmes1998} algorithm. It
finds iteratively the partition in $k$ disjoint subsets that maximizes the
likelihood of the corresponding Gaussian mixture model. The choice of the
$k$ is done online as follows. Let $C_k = \{c_1, \dots, c_k\}$ be
the clustering into $k$ disjoint subsets obtained by the Expectation-Maximization
algorithm. Let $|c|$ and $\varpos(c)$ denote respectively the number of
particles and the variance of the particles positions. Finally, we define
respectively the internal variances of position and orientation of $C_k$ as
\[
  \varpos(C_k) = \frac{
    \sum_{i=0}^k \varpos(c_i) * |c_i|
  }{
    \sum_{i=0}^k{|c_i|}
  }.
\]
Starting from $k=1$, the number of clusters is increased until we reach $k=5$
or $\varpos(C_k) \geq 0.5 \varpos(C_{k+1})$.
This ensures that adding a cluster is only done if it provides a major reduction
of the variance.
When several clusters are considered, only the most populated one is considered
for high-level decisions.

\begin{figure}[h!]
  \subfloat[Filtering of scattered particles: particles in blue, best cluster mean in red.]{
    \includegraphics[width=.495\textwidth]{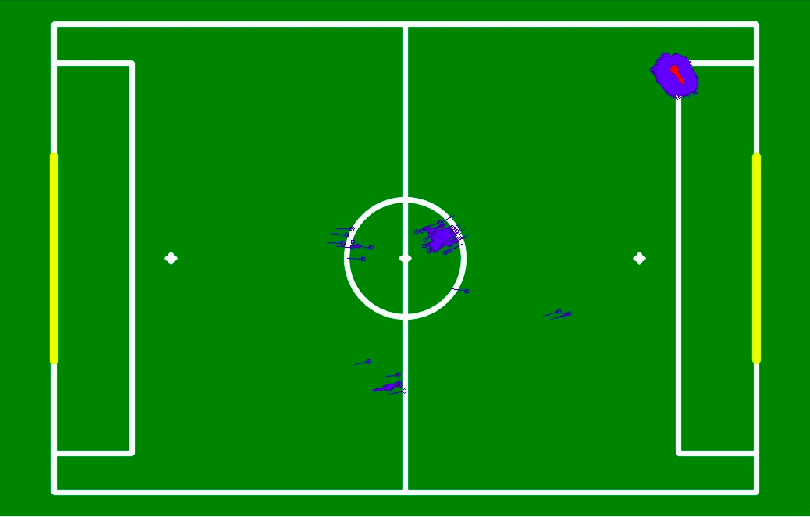}
    \label{sub:locmatch}
  }
  \hfill
  \subfloat[Monitoring on field entry: Robot 1 enters the field from the side line.]{
    \includegraphics[width=.45\textwidth]{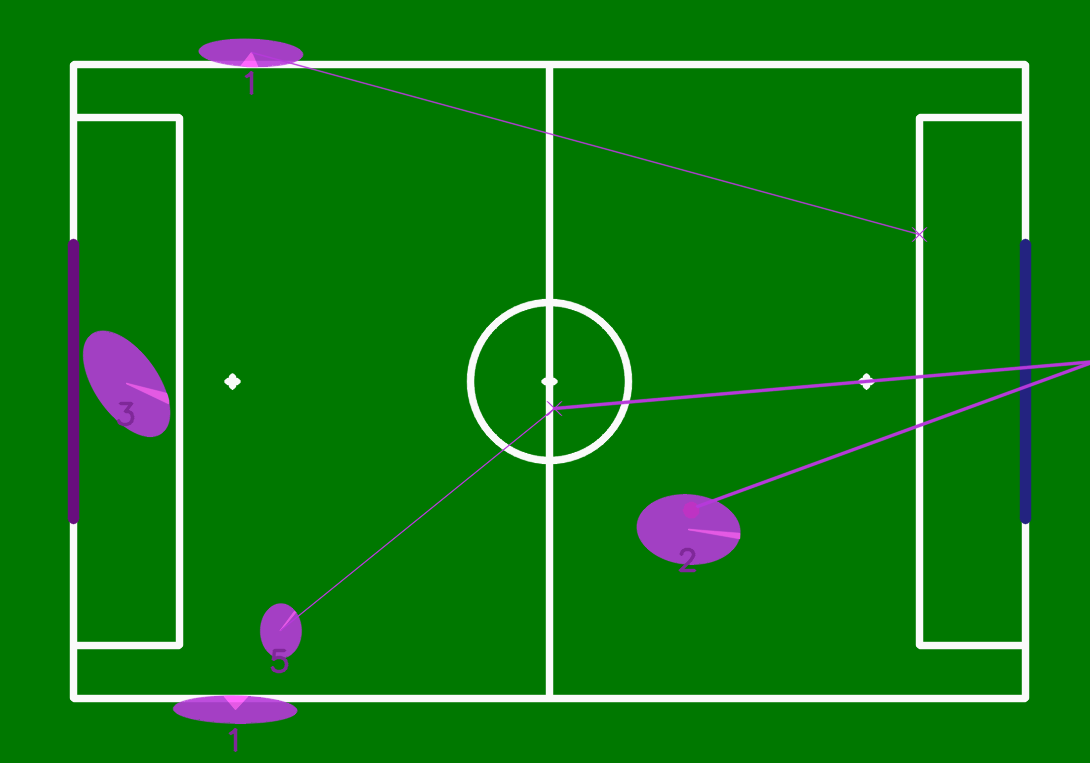}
    \label{sub:monitoring_border}
  }
  \caption{In-game examples of localization.}
  \label{fig:loc}
\end{figure}

The proposed method provides two main benefits with respect to the simple
solution of taking the average of all particles. First, it allows to obtain
meaningful orientation while the particle filter has not converged. This
situations frequently occurs when a robot comes back from penalty from the side
line. During games, it allows to discard scattered particles which will be
automatically attributed to a cluster containing the noise, see
Fig.~\ref{sub:locmatch}. Second, representation of localization belief under
the form of multiple clusters allows to store or broadcast the information at a
much lower cost than sending the position of all the particles. Therefore, this
method enables real-time monitoring under low bandwidth conditions, see
Fig.~\ref{sub:monitoring_border}
\section{\label{sec:strategy}Strategy}
In 2017, we introduced an off-line value iteration process to compute a kick
policy that chooses a type of kick as well as the orientation aiming to optimize
the time to score a goal with one robot on an empty field\cite{n2018rhoban}. The
reward function used for the optimization is simply the time needed for the
robot to reach the next ball position, a penalty if the ball is kicked out of
field and 0 if a goal is scored. This process produces a value function $V$ that
gives us an estimation of the time it takes to score a goal from a given
position on the field.

This year, we also used an online value iteration that performs an optimization
at depth one to include the current state of the game. It can roughly be
described as follows. The discrete set of ball positions on the field is called
$S$. An action is a type of kick together with a discrete orientation. We denote
$A$ the set of possible actions. Let $P_a(s, s')$ denotes the probability of
reaching state $s'$ from state $s$ after performing action a kick with
orientation $a$. The knowledge of the game is introduced with a reward function
$r(s,s)$ described thereafter. The online policy is given by the formula
\[
    \arg\max_{a \in A} \sum_{s' \in S} P_a(s, s') (r(s, s') + V(s')).
\]
To compute the reward $r$, we check if we are in a state where it is not allowed
to score a goal. For example in the case of a throw-in, an indirect penalty kick
or when we have the kick-off and the ball is still not in play (exited the
center circle). In this case, a penalty score is given if a goal is scored. In
that case, the robot will naturally kick the ball so that it does not score a
goal, but placing it in the best situation possible to score a goal on the next
kick. The reward $r(s,s')$ also includes the time for the closest robot to reach
$s'$ from which we subtract the time for the kicking robot to reach $s'$.
Finally, we give penalties to kicks towards opponent robots if we have their
location.
\begin{figure}[htb]
  \centering
  \includegraphics[width=0.45\linewidth]{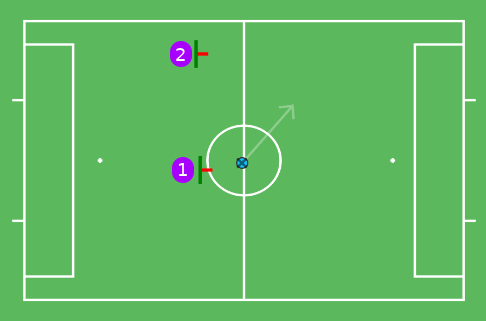}
  \caption{An example situation using on-line Monte Carlo.}
  \label{fig:montecarlo}
\end{figure}
As an example, in the situation of Fig.~\ref{fig:montecarlo}, the
robot 1 is going for the ball, because he is the closest to it. He is not
allowed to score a goal, because of the kick-off conditions. Hence, a short kick
action getting the ball out of the center circle is preferred by the on-line
iteration instead of a powerful kick that would certainly score. If the robot
was alone on the field, the optimal orientation would be straight forward.
However, the iteration produces this left oriented kick, because by the time we
estimate the kick is done, the robot 2 will be properly positioned to handle
the ball and score a goal faster than if robot 1 would have kicked straight.
%
%
%
\bibliographystyle{splncs04}
\bibliography{ChampionPaper2019}
\end{document}